\documentclass{ceurart}

\usepackage{booktabs}
\usepackage{amssymb}
\usepackage{algorithm}
\usepackage{algpseudocode}

\newcommand{\ies}[1]{\texttt{ies:#1}}
\newcommand{\hqd}[1]{\texttt{hqdm:#1}}
\newcommand{\bfo}[1]{\texttt{bfo:#1}}
\newcommand{\sub}{\ensuremath{\sqsubseteq}}
\newcommand{\lra}{\texorpdfstring{\ensuremath{\leftrightarrow}}{<->}}
\newcommand{\equivc}{\ensuremath{\equiv}}

\begin{document}

\copyrightyear{2026}
\copyrightclause{Copyright for this paper by its authors.
  Use permitted under Creative Commons License Attribution 4.0
  International (CC BY 4.0).}

\conference{OM-2026: The 21st International Workshop on Ontology Matching,
  collocated with the 25th International Semantic Web Conference ISWC-2026,
  October 25th or 26th, 2026, Bari, Italy}

\title{Consistency Is Not Coherence: Orientation Search for Certified
  Alignments Between 4D Defence Upper Ontologies}

\author[1]{Fabio Rovai}[%
orcid=0009-0001-7244-794X,
email=fabio@thetesseractacademy.com,
url=https://gov.tesseract.academy,
]
\address[1]{The Tesseract Academy (Kampakis and Co Ltd), 5 Brunswick Park
  Gardens, London N11 1EJ, United Kingdom}

\begin{abstract}
  We align three upper ontologies that sit under UK and NATO defence data
  infrastructure: the Information Exchange Standard (IES), the Higher Quality
  Data Model (HQDM) that underpins the National Digital Twin, and Basic Formal
  Ontology (BFO). No public alignment between IES and HQDM existed. Promoting a
  hand-curated 17-correspondence crosswalk to OWL and reasoning over the
  complete merged ontologies with HermiT produces three results that we believe
  matter beyond this pair. First, the published \texttt{hqdm.owl} shipped by
  GCHQ is \emph{natively incoherent}: 39 of its 229 named classes are
  unsatisfiable before any alignment is added, while IES and BFO have none. An
  alignment evaluated against it inherits a broken target. Second, naive
  equivalence promotion leaves the merge \emph{consistent}, so the usual check
  passes, while creating 100 new unsatisfiable classes and 218 entailed
  source-internal subsumptions that neither standard asserts. Consistency is the
  wrong acceptance test for an alignment. Third, we present \emph{orientation
  search}, a repair operator that treats the direction of each correspondence
  ($\sqsubseteq$, $\sqsupseteq$, $\equiv$, discard) as the search variable
  rather than treating mapping deletion as the only move, with a reasoner as
  oracle for coherence and conservativity. It yields a 21-axiom IES\lra{}HQDM
  bridge with zero new unsatisfiable classes and zero conservativity violations,
  in which every weakening carries a machine-found counterexample as its
  justification. Crossing to BFO is possible only through the occurrent branch:
  with the rest of the bridge fixed, mapping \ies{Entity} to
  \bfo{material~entity} empties 101 IES classes, weakening that mapping to a
  subsumption still empties 100, and re-targeting it to \bfo{history} empties
  none. The defect is the target, not the strength. Finally, we test whether existing systems produce this bridge.
  LogMap and LogMapLt accept the crosswalk's documented false friend
  \ies{Event}~\equivc{}~\hqd{event}, map onto natively unsatisfiable classes,
  and carry 87 and 214 new unsatisfiable classes once read as logic; LogMap's
  repair, given our expert input, is safe but keeps none of the five provable
  equivalences; two LLM oracles score 6/12 and 7/12. Every system, classical
  and neural, falls for the same false friend the reasoner-refereed search
  avoids. All artifacts are open.
\end{abstract}


\maketitle

\section{Introduction}
\label{sec:intro}

Alignments are usually published as sets of correspondences with a similarity
score, and consumed as if they were logical axioms. The gap between those two
readings is where this paper lives.

Our setting is deliberately unglamorous and operationally real. The UK
Information Exchange Standard (IES) is the vocabulary in which UK defence and
national-security systems exchange assertions about people, events and places.
The Higher Quality Data Model (HQDM)~\citep{west2011} is the foundation data
model beneath the National Digital Twin and the built-environment estate. Both
descend from the same BORO and ISO 15926 4D tradition~\citep{partridge2005}, so
they ought to be cousins, and an autonomous system reasoning about a mission in
a real place has to join what IES says about the operational picture to what
HQDM says about the terrain. Basic Formal Ontology (BFO)~\citep{arp2015,iso21838}
is the third party in the room because it is the realist upper ontology of the
US and NATO defence data foundry, so a UK-to-coalition join runs through it. No
public machine-readable alignment between IES and HQDM existed before this work.
For IES and BFO, a thirteen-author team spanning the IES, BORO and BFO
communities has recently compared the two frameworks at the level of design
patterns~\citep{bailey2025comparing}, explicitly deferring ``axiomatic
translation definitions'' that would move ``from plausible semantic equivalence
to provable equivalence'' to future work. This paper supplies that layer:
every correspondence here is a reasoner-certified OWL axiom.

We built one, then asked the question that the ontology matching community has
long treated as central and that alignment consumers still routinely skip: what
happens when the correspondences are read as logic? The answer, on real
published artifacts, is worse than the folklore suggests, and the repair is more
interesting than deletion.

\paragraph{Contributions.}
\begin{enumerate}
\item \textbf{A native-defect finding} (\S\ref{sec:native}). The published
  \texttt{hqdm.owl} has 39 unsatisfiable classes with no alignment present,
  invisible both in the asserted hierarchy and to a consistency check. Any
  evaluation of an alignment \emph{into} HQDM measures against a broken target,
  and any mapping onto one of the 39 is vacuously true and semantically empty.
\item \textbf{A separation result on real data} (\S\ref{sec:naive}). Naive
  equivalence promotion of a 17-correspondence expert crosswalk yields a merge
  that is consistent, and therefore passes the check most pipelines run, while
  adding 100 new unsatisfiable classes and 218 conservativity violations. We give
  the isolation: exactly one of the ten class targets is natively unsatisfiable,
  and weakening exactly that one edge takes new unsatisfiability from 100 to 0.
  One defective source axiom, one correspondence, one hundred dead operational
  classes.
\item \textbf{Orientation search} (\S\ref{sec:method}), a repair operator in
  which the search variable is the \emph{direction} of each correspondence rather
  than its presence. Standard alignment repair removes mappings from a diagnosis;
  orientation search first weakens $\equiv$ to a subsumption, then attempts
  completion back toward $\equiv$, with the reasoner refereeing every step against
  a coherence criterion and a conservativity criterion. Each rejected direction is
  returned with the reasoner's counterexample, so the artifact documents the
  disagreement between the two standards rather than hiding it.
\item \textbf{Certified bridges} (\S\ref{sec:results}): a 21-axiom
  IES\lra{}HQDM bridge produced by the search, an IES\lra{}BFO bridge whose four
  targets were specified by hand and then certified, and their union over the
  complete three ontologies with direct HQDM\lra{}BFO anchors, all at zero new
  unsatisfiable classes and zero conservativity violations. An ablation on the
  BFO bridge (\S\ref{sec:bfo}) separates a mapping that is too strong from one
  that is aimed at the wrong target, and shows that only the second diagnosis is
  the right one here.
\item \textbf{A systems comparison} (\S\ref{sec:systems}). LogMap and LogMapLt,
  evaluated on the same pairs under the same two criteria, accept the documented
  false friend, map onto natively unsatisfiable classes, and carry 87 and 214
  new unsatisfiable classes plus over a hundred conservativity violations each.
  Given our expert input, LogMap's repair returns a certifiably safe alignment,
  but a uniformly reversed one that loses all five provable equivalences. An
  LLM-oracle probe (6/12 base, 7/12 IES fine-tune) completes the picture:
  classical and neural systems all fail on the same pairs, the ones where
  lexical evidence misleads, which is where the reasoner is doing work no other
  component substitutes for.
\end{enumerate}

All ontologies used are public, all code and certified artifacts are released,
and the reasoning is deterministic (see the availability statement at the end
of the paper).

\section{Materials and background}
\label{sec:background}

\subsection{The three ontologies}

Table~\ref{tab:onts} reports what we measured on the exact vendored files, not
what the documentation claims. Two properties are worth noting immediately. IES
declares 511 named classes and \emph{zero} class-disjointness axioms; it is a
rich taxonomy with almost no exclusion structure. HQDM declares 229 named
classes and 14 disjointness axioms, introduced by the EXPRESS-to-OWL rendering.
BFO 2020 is small (35 named classes) and heavily constrained (19 disjointness
axioms), which is exactly why it is the ontology that punishes a careless
mapping hardest.

\begin{table}[h]
\centering
\caption{The three ontologies as measured on the vendored files. Named classes
counts \texttt{rdf:type owl:Class} subjects with an IRI. Disjointness counts
\texttt{owl:disjointWith} triples; none of the three uses
\texttt{owl:AllDisjointClasses} or \texttt{owl:disjointUnionOf}. Unsatisfiable
classes are computed by HermiT on each ontology alone.}
\label{tab:onts}
\begin{tabular}{lrrrrr}
\toprule
Ontology & triples & named classes & object props & disjointness & unsat (alone) \\
\midrule
IES (\texttt{ies-common.ttl})  & 4{,}039 & 511 & 162 & 0  & \textbf{0} \\
HQDM (\texttt{hqdm.owl})       & 3{,}127 & 229 & 74  & 14 & \textbf{39} \\
BFO 2020 (\texttt{bfo.owl})    & 1{,}221 & 35  & 0   & 19 & \textbf{0} \\
\bottomrule
\end{tabular}
\end{table}

A note on names: the \texttt{dstl/IES4} repository was archived in March 2025
(custodianship moved from Dstl to the cross-government IES Working Group under
DBT; last public release v4.3.1); the canonical successor is
\texttt{IES-Org/ont-ies}, whose \texttt{ies-common.ttl} we vendor and measure.
Counts differ between lineages, so all IES claims here are about the file in
Table~\ref{tab:onts}, identified by content hash in the released artifact.

IES and HQDM are both four-dimensionalist: an individual is a spatio-temporal
extent, and what a three-dimensionalist calls a state of an object is a temporal
part of that extent. BFO is three-dimensionalist at its core, partitioning
existence into \bfo{continuant} and \bfo{occurrent} and declaring them disjoint.
That single disjointness is the whole difficulty of the IES-to-BFO join, and
\S\ref{sec:bfo} shows it is crossable but only in one place.

\subsection{The crosswalk}

The input alignment is a hand-curated crosswalk of 17 correspondences in
SSSOM~\citep{sssom}: 13 \texttt{skos:closeMatch} (10 class pairs, 3 property
pairs) and 4 \texttt{skos:relatedMatch} recorded as warnings rather than as
mappings. It is accompanied by six documented divergences, pairs that look like
they map and do not. The headline divergence is a false friend that any
label-based matcher gets exactly backwards: \ies{Event} is a durative happening
with participants, while \hqd{event} is an instantaneous temporal boundary with
no duration and no participants. The true counterpart of \ies{Event} is
\hqd{activity}. The crosswalk records \ies{Event}~$\sim$~\hqd{event} at
confidence 0.25 purely to carry the warning. We return to this pair in
\S\ref{sec:llm}, where both language models we test walk straight into it.

\subsection{Consistency, coherence, conservativity}

We use three standard criteria and are pedantic about the difference because the
central negative result of this paper is that the first one is useless here.

An ontology $\mathcal{O}$ is \emph{consistent} if it has a model, and
\emph{coherent} if no named class is unsatisfiable ($\mathcal{O} \not\models
C \sqsubseteq \bot$ for named $C$). Coherence implies consistency; the
converse fails, and not exotically: a consistent ontology can be satisfied by a
model in which a hundred of its classes are simply empty. Alignment incoherence
and its repair are well
studied~\citep{meilicke2011,meilicke2009,jimenez2011logmap}, and yet
consistency checking remains what deployment pipelines actually run.

The third criterion is \emph{conservativity}~\citep{solimando2014,cuencagrau2008}.
Let $\mathcal{O}_1, \mathcal{O}_2$ be the input ontologies and $\mathcal{M}$ the
alignment read as axioms. $\mathcal{M}$ violates conservativity if the merge
entails a subsumption between two classes of the \emph{same} input ontology that
the input ontology did not itself entail. Such an entailment is a statement about
IES that IES never made, produced by the act of aligning it to something else. A
consumer who trusts the merged model is now reasoning with axioms no standards
body ever wrote.

Our acceptance criteria are therefore:
\begin{description}
\item[S1$'$ (relative coherence).] The merge introduces no unsatisfiable class
  beyond those already unsatisfiable in the inputs alone. We use the relative
  form because HQDM's 39 native casualties (\S\ref{sec:native}) would otherwise
  make every candidate fail identically and carry no signal.
\item[S2 (conservativity).] The merge entails no new subsumption between two
  classes of the same input ontology.
\end{description}

\section{Orientation search}
\label{sec:method}

\subsection{The operator}

Alignment repair conventionally computes a diagnosis, a minimal subset of
mappings whose removal restores coherence, and deletes it. Deletion is a blunt
move: it throws away a correspondence that a domain expert asserted for good
reason, and it records nothing about \emph{why} the correspondence failed.

Orientation search replaces deletion with re-orientation. A curated crosswalk
gives us $(s, o)$ pairs but, because SKOS carries no logical commitment, it does
not tell us the logical strength of the relation. That strength is precisely
what we should be searching over. For each correspondence we consider the
lattice
\[
\{\, s \equiv o \,\} \;\succ\; \{\, s \sqsubseteq o \,\},\ \{\, s \sqsupseteq o \,\}
\;\succ\; \{\, \emptyset \,\}
\]
and, when even the forward subsumption fails because the target is defective, a
fifth move: \emph{weaken to ancestor}, replacing $o$ by a superclass of $o$,
in the spirit of axiom weakening~\citep{troquard2018}.

The search runs in two phases. \textbf{P1} accepts each forward edge
$s \sqsubseteq o$, weakening the target upward while the edge fails S1$'$ or
S2. \textbf{P2} greedily attempts each reverse edge $o \sqsubseteq s$; where
it survives, the pair completes to a full equivalence. The result declares the
standards equivalent exactly where they agree and one-directionally linked
exactly where they do not.

\begin{algorithm}[h]
\caption{Orientation search}
\label{alg:orient}
\begin{algorithmic}[1]
\Require input ontologies $\mathcal{O}_1,\mathcal{O}_2$; correspondences $C$; reasoner $\mathcal{R}$
\State $U_{\mathrm{nat}} \gets \mathcal{R}.\textsc{Unsat}(\mathcal{O}_1 \cup \mathcal{O}_2)$ \Comment{native baseline}
\State $B \gets \emptyset$
\ForAll{$(s,o) \in C$} \Comment{P1: forward, weakening to ancestor}
  \State $t \gets o$
  \While{$t \neq \top$ \textbf{and} $\lnot\textsc{Certify}(B \cup \{s \sqsubseteq t\})$}
    \State $t \gets \textsc{Parent}(t)$
  \EndWhile
  \If{$t \neq \top$} $B \gets B \cup \{s \sqsubseteq t\}$ \EndIf
\EndFor
\ForAll{$(s,o) \in C$ with $s \sqsubseteq o \in B$} \Comment{P2: completion toward $\equiv$}
  \If{$\textsc{Certify}(B \cup \{o \sqsubseteq s\})$} $B \gets B \cup \{o \sqsubseteq s\}$
  \Else{} \textbf{record} $\mathcal{R}$'s witness as the justification for one-directionality
  \EndIf
\EndFor
\State \Return $B$
\Statex
\Function{Certify}{$B$}
  \State $U \gets \mathcal{R}.\textsc{Unsat}(\mathcal{O}_1 \cup \mathcal{O}_2 \cup B)$
  \State $V \gets \{\, C_1 \sqsubseteq C_2 \ \text{entailed by the merge} \mid C_1, C_2 \in \mathcal{O}_i,\ \mathcal{O}_i \not\models C_1 \sqsubseteq C_2 \,\}$
  \State \Return $(U \setminus U_{\mathrm{nat}} = \emptyset) \wedge (V = \emptyset)$ \Comment{S1$'$ and S2}
\EndFunction
\end{algorithmic}
\end{algorithm}

\subsection{Instantiation}

The reasoner is HermiT~\citep{glimm2014} via owlready2, called on the complete
merged ontologies, not on fragments and not on hand-selected axiom subsets. The
run costs on the order of 35 reasoner calls and completes in about two minutes
on a laptop. The procedure is deterministic: HermiT is a DL prover with no
learned component, so the same inputs give the same certified output.

Two design decisions are worth defending. First, S1$'$ is relative rather than
absolute. If we demanded absolute coherence, every candidate would fail on
HQDM's 39 native casualties and the search would return nothing; the relative
form isolates the damage the \emph{alignment} does. Second, P2 is greedy rather
than exhaustive, so the returned bridge is a maximal certified bridge under a
fixed correspondence order rather than a provably maximum one. We consider this
an honest limitation (\S\ref{sec:limits}) rather than a defect, since every
axiom in the output is individually certified regardless of order.

\section{Results}
\label{sec:results}

\subsection{The target ontology is broken before we touch it}
\label{sec:native}

Reasoned alone, \texttt{hqdm.owl} has \textbf{39 unsatisfiable classes}
(Table~\ref{tab:onts}). IES alone and BFO alone have none. The casualties are
concentrated in HQDM's relationship-derived branch, including \hqd{association},
\hqd{employment}, \hqd{ownership}, \hqd{participant}, \hqd{sign},
\hqd{asset}, \hqd{transfer\_of\_ownership} and \hqd{physical\_quantity}. The
mechanism is the collision between the EXPRESS relationship heritage, in which a
relationship is a first-class entity, and the \hqd{class}/\hqd{relationship} and
\hqd{abstract\_object}/\hqd{spatio\_temporal\_extent} disjointness axioms
introduced by the OWL rendering. The file's own header warns of known
consistency issues; what we add is the quantification and, more importantly, the
identification of \emph{which} classes are affected, because that is what an
alignment consumer needs.

Three consequences follow for ontology matching practice. A matcher that maps
onto any of those 39 classes has produced a correspondence whose target is
provably empty: the mapping is vacuously satisfied and carries no information. A
reference alignment built by human curators against this file may contain such
mappings, since the defect is invisible in the asserted hierarchy. And an
evaluation that reports precision and recall against such a reference is
reporting agreement about empty sets. We suggest that a coherence report on each
\emph{input} ontology should be a routine part of publishing a matching test
case, alongside class and property counts.

\subsection{Consistent, and thoroughly incoherent}
\label{sec:naive}

We promote all 13 \texttt{closeMatch} correspondences to \texttt{owl:equivalentClass}
and \texttt{owl:equivalentProperty} and merge the complete ontologies. The result:

\begin{table}[h]
\centering
\caption{Naive equivalence promotion of the expert crosswalk. The consistency
check, which is what deployment pipelines typically run, passes.}
\label{tab:naive}
\begin{tabular}{lr}
\toprule
Criterion & Result \\
\midrule
Consistency (a model exists) & \textbf{passes} \\
New unsatisfiable classes (beyond the native 39) & \textbf{100} \\
New source-internal subsumptions (S2 violations) & \textbf{218} \\
\bottomrule
\end{tabular}
\end{table}

The 100 casualties are IES operational classes: \ies{Accused}, \ies{Arrested},
\ies{ArrestingOfficer}, \ies{Assessor}, \ies{Witness}, \ies{Customer},
\ies{Authoriser}, \ies{AtWar}. These are the classes an intelligence or policing
application actually instantiates. The 218 conservativity violations include
\ies{Entity} \sub{} \ies{State}, \ies{Investigation} \sub{} \ies{State} and
\ies{OfferForSale} \sub{} \ies{Entity}: statements about IES, in IES's own
vocabulary, that IES does not make and that arise solely from having been
aligned to HQDM.

\paragraph{Isolation.} Exactly one of the ten class targets, \hqd{participant},
is among HQDM's 39 native casualties. The certified run (\S\ref{sec:bridge})
weakens exactly one edge, \ies{EventParticipant} \sub{} \hqd{participant}
replaced by \ies{EventParticipant} \sub{} \hqd{state}, one hop up the hierarchy,
and new unsatisfiability falls from 100 to 0. The single edge into the defective
region is therefore what wires IES into HQDM's broken branch, and one
correspondence onto one bad class takes down a hundred classes in the other
ontology. This is the strongest practical argument we can make for reporting
input-ontology coherence: the cost of a single mapping onto an unsatisfiable
target is not local.

\subsection{The certified IES\lra{}HQDM bridge}
\label{sec:bridge}

Orientation search returns \textbf{21 directed axioms} at 0 new unsatisfiable
classes and 0 conservativity violations (Table~\ref{tab:bridge}). Five class
pairs and all three property pairs complete to full equivalence. Four class
pairs remain one-directional, each carrying the reasoner's counterexample, and
one is weakened.

\begin{table}[h]
\centering
\caption{The certified bridge. Prefixes are omitted: every source is
\texttt{ies:} and every target \texttt{hqdm:}. The last column is the number of
S2 violations the \emph{reverse} edge would have introduced, which is why the
row is one-directional. Every verdict is a reasoner decision, not a curator's
judgement.}
\label{tab:bridge}
\small
\begin{tabular}{lllr}
\toprule
Source (\texttt{ies:}) & Target (\texttt{hqdm:}) & Verdict & S2 if reversed \\
\midrule
\texttt{Element}          & \texttt{spatio\_temporal\_extent}          & full \equivc{} & 0 \\
\texttt{Thing}            & \texttt{thing}                             & full \equivc{} & 0 \\
\texttt{Event}            & \texttt{activity}                          & full \equivc{} & 0 \\
\texttt{PossibleWorld}    & \texttt{possible\_world}                   & full \equivc{} & 0 \\
\texttt{ClassOfElement}   & \texttt{class\_of\_spatio\_temporal\_extent} & full \equivc{} & 0 \\
\midrule
\texttt{Entity}           & \texttt{individual}     & forward only & 104 \\
\texttt{State}            & \texttt{state}          & forward only & 109 \\
\texttt{PeriodOfTime}     & \texttt{period\_of\_time} & forward only & 1 \\
\texttt{ParticularPeriod} & \texttt{period\_of\_time} & forward only & 5 \\
\texttt{EventParticipant} & \texttt{participant} $\to$ \texttt{state} & weakened & n/a \\
\midrule
\texttt{isPartOf}         & \texttt{part\_of}           & full \equivc{} & 0 \\
\texttt{isParticipantIn}  & \texttt{participant\_in}    & full \equivc{} & 0 \\
\texttt{isStateOf}        & \texttt{temporal\_part\_of} & full \equivc{} & 0 \\
\bottomrule
\end{tabular}
\end{table}

The witnesses the reasoner returned for the four rejected reverse edges are the
scientific content of this result, so we give them in full. Each violation set
is stored sorted, and every witness quoted below is among the first six entries
of its sorted set, so the citations are reproducible rather than an artifact of
the order in which the reasoner happened to enumerate them:

\begin{description}
\item[\ies{Entity} $\sim$ \hqd{individual} (104 violations).] The reverse forces
  \ies{Arrest} \sub{} \ies{Entity}, and likewise \ies{AccountAdminEvent},
  \ies{AgreementExecution}, \ies{Assessment} and 100 further IES classes. HQDM
  asserts \hqd{activity} \sub{} \hqd{individual}, while IES holds \ies{Event}
  and \ies{Entity} apart, so every IES happening is dragged under \ies{Entity}.
  HQDM's \hqd{individual} is strictly broader than IES's notion of a whole-life
  persisting thing.
\item[\ies{State} $\sim$ \hqd{state} (109 violations).] The reverse forces
  \ies{Arrest} \sub{} \ies{State} by the same route, since HQDM's
  \hqd{individual} sits under \hqd{state}. An arrest becomes a state of IES.
\item[\ies{PeriodOfTime} $\sim$ \hqd{period\_of\_time} (1 violation).] The
  reverse forces \ies{PossibleWorld} \sub{} \ies{PeriodOfTime}, because HQDM
  asserts \hqd{possible\_world} \sub{} \hqd{period\_of\_time} and IES does not.
\item[\ies{ParticularPeriod} $\sim$ \hqd{period\_of\_time} (5 violations).] The
  reverse forces \ies{PeriodOfTime} \sub{} \ies{ParticularPeriod}, collapsing
  the distinction between a period and an ISO 8601-anchored particular one, and
  likewise for \ies{ArbitraryPeriod}, \ies{RecurringPeriod} and
  \ies{PossibleWorld}.
\end{description}

These are not failures of the crosswalk. They are the two standards disagreeing,
localised to a named class by a prover. The \ies{Entity}~$\sim$~\hqd{individual}
case says something a human curator can act on: HQDM subsumes activities under
individuals, IES insists an event is not an entity, so the two notions of
whole-life persisting thing are genuinely different and the correct relation is
a subsumption, not an identity. That is a finding about the standards produced
as a by-product of certifying the alignment, and it is exactly the kind of
content that an alignment published as a bare similarity score cannot carry.

We note the shape of the outcome. Where the two 4D models genuinely agree, the
bridge is full equivalence. Where they disagree, it is exactly as weak as the
disagreement requires and no weaker. A diagnosis-and-delete repair would have
returned a smaller alignment with less information in it.

\subsection{Crossing to BFO: only via the occurrent branch}
\label{sec:bfo}

BFO is where the 4D/3D mismatch has to be paid for. Unlike the HQDM case, the
four BFO targets were specified by hand and then certified; orientation search
verified and oriented them but did not discover them, and \S\ref{sec:limits}
explains why weakening alone could not have.

The bridge is

\begin{center}
\begin{tabular}{ll}
\ies{Element} \sub{} \bfo{occurrent} & \ies{Entity} \sub{} \bfo{history} \\
\ies{Event} \sub{} \bfo{process}     & \ies{State} \sub{} \bfo{occurrent} \\
\multicolumn{2}{l}{\texttt{cwb:hasHistory} : \bfo{material~entity} $\rightarrow$ \bfo{history} (declared by the bridge)} \\
\end{tabular}
\end{center}

certified at 0 new unsatisfiable classes and 0 conservativity violations. The
one axiom in it that a curator would not write is \ies{Entity} \sub{}
\bfo{history}. The obvious choice is \ies{Entity} \equivc{}
\bfo{material~entity}, and Table~\ref{tab:bfo} shows what that choice costs by
changing \emph{only} that axiom and holding the other three fixed.

\begin{table}[h]
\centering
\caption{The cost of one axiom. Every row merges the complete IES and BFO with
\ies{Element} \sub{} \bfo{occurrent}, \ies{Event} \sub{} \bfo{process} and
\ies{State} \sub{} \bfo{occurrent} held fixed, and varies only how \ies{Entity}
is mapped. Counts are IES classes made unsatisfiable.}
\label{tab:bfo}
\begin{tabular}{llr}
\toprule
Mapping for \ies{Entity} & Rest of bridge & IES classes collapsed \\
\midrule
\equivc{} \bfo{material~entity} & absent  & 0 \\
\sub{} \bfo{material~entity}    & absent  & 0 \\
\equivc{} \bfo{material~entity} & present & \textbf{101} \\
\sub{} \bfo{material~entity}    & present & \textbf{100} \\
\sub{} \bfo{history}            & present & \textbf{0} \\
\bottomrule
\end{tabular}
\end{table}

Three things follow. First, the collapse is a property of the bridge as a
whole, not of one mapping in isolation: aligning \ies{Entity} to
\bfo{material~entity} on its own is harmless, because IES declares no
disjointness and nothing then connects it to BFO's occurrent branch. Only once
the rest of the bridge puts IES's 4D backbone on that branch does the
\bfo{continuant}/\bfo{occurrent} disjointness fire and empty 101 IES classes
(\ies{Account}, \ies{Actor}, \ies{Aircraft}, \ies{Bank}, \ldots). An alignment
cannot be certified one correspondence at a time.

Second, weakening does not rescue it. Replacing the equivalence with a plain
subsumption still kills 100 classes. This is the case that shows why the
weaken-to-ancestor operator of \S\ref{sec:method} is not sufficient on its own:
walking up from \bfo{material~entity} reaches \bfo{independent~continuant},
\bfo{continuant} and finally \bfo{entity}, at which point the axiom is safe but
vacuous. The ancestor chain never passes through \bfo{history}.

Third, re-targeting does rescue it, completely. BFO 2020 already ships the class
a 4D standard is talking about: \bfo{history} (\texttt{BFO\_0000182}), a process
that is the totality of what happens to a material entity, sitting natively on
the occurrent branch. Changing that single axiom takes the count from 101 to 0.
The naive mapping was not too strong; it was aimed at the wrong target, and the
distinction matters because only one of those two diagnoses is fixed by
weakening. A 4D \ies{Entity} is not a BFO continuant; it is the continuant's
history. The bridge declares \texttt{cwb:hasHistory} so that a consumer can
still get from the continuant to the 4D object, which is the relation the naive
equivalence was groping for.

\subsection{All three in one artifact}

The IES\lra{}HQDM and IES\lra{}BFO bridges are not merely
individually safe. Their union over the complete IES $\cup$ HQDM $\cup$ BFO is
also certified at 0 new unsatisfiable classes and 0 conservativity violations,
and it admits direct HQDM\lra{}BFO anchors on top, found the same
reasoner-refereed way:
\hqd{spatio\_temporal\_extent} \equivc{} \bfo{occurrent},
\hqd{individual} \equivc{} \bfo{history},
\hqd{activity} \sub{} \bfo{process},
\hqd{state} \sub{} \bfo{occurrent},
\bfo{temporal~region} \sub{} \hqd{period\_of\_time}.
The fused artifact is 32 subsumption axioms plus the \texttt{cwb:hasHistory}
declaration. \hqd{participant} is deliberately excluded from the
HQDM\lra{}BFO anchors: it is one of the 39 native casualties, so any
mapping onto it would be vacuously true.

That the two independently certified bridges compose without new violations
is not guaranteed by their individual certificates; we checked it rather than
assumed it. Composition of certified alignments must, in general, be
re-certified.

\section{Do existing systems produce this bridge?}
\label{sec:systems}

Mature systems exist for both halves of this problem: matchers that propose
correspondences, and repair facilities that restore coherence. We ran the
natural incumbents on exactly this material and evaluated every output with the
same HermiT harness and criteria. Three questions: does a state-of-the-art
matcher \emph{find} the expert backbone; is its output \emph{safe} to consume
as logic; and, given the expert crosswalk, does an established \emph{repair}
facility recover what orientation search recovers?

\subsection{Classical matchers on the pair}
\label{sec:matchers}

We ran LogMap and LogMapLt (the July-2021 standalone distribution,
\texttt{logmap-matcher-4.0.jar}, default configuration) on the vendored
IES--HQDM pair, and LogMap on IES--BFO. Table~\ref{tab:systems} evaluates each
output alignment, read as axioms per the system's own asserted relations,
against S1$'$ and S2.

\begin{table}[h]
\centering
\caption{Matcher outputs on the vendored pairs, evaluated with HermiT under the
same criteria as the certified bridge. ``Backbone'' scores correct target /
wrong target / missed against the pair's gold targets: the ten expert class
correspondences for IES--HQDM, the four certified bridge targets for IES--BFO.
``Unsat targets'' are accepted mappings whose HQDM target is one of the 39
natively unsatisfiable classes. The certified bridge row is the output of
\S\ref{sec:bridge}.}
\label{tab:systems}
\small
\begin{tabular}{lrrrrrr}
\toprule
System (pair) & mappings & backbone & false friend & unsat targets & new unsat & S2 viol. \\
\midrule
LogMap (IES--HQDM)   & 24 & 3/1/6  & \textbf{yes} & 2 & \textbf{87}  & \textbf{103} \\
LogMapLt (IES--HQDM) & 36 & 3/1/6  & \textbf{yes} & 2 & \textbf{214} & \textbf{112} \\
LogMap (IES--BFO)    & 2  & 0/1/3  & ---          & 0 & 0            & \textbf{2}   \\
\midrule
certified bridge (IES--HQDM) & 13 & 10/0/0 & no & 0 & \textbf{0} & \textbf{0} \\
\bottomrule
\end{tabular}
\end{table}

Three observations, each of which independently justifies a reasoner in the
loop.

First, \textbf{both matchers accept the false friend}
\ies{Event}~\equivc{}~\hqd{event}, and neither finds
\ies{Event}~$\sim$~\hqd{activity}. The consequence is not merely a wrong pair:
read as logic, LogMap's output entails \hqd{activity} \sub{} \hqd{event}
\emph{inside HQDM} (every durative activity becomes an instantaneous
boundary): an S2 violation against an ontology the matcher was supposed to
align, not rewrite. Both matchers also accept mappings onto \hqd{asset} and
\hqd{agreement\_execution}, two of the 39 natively unsatisfiable classes
(vacuously true correspondences), and both map RDF(S) vocabulary itself
(\texttt{rdfs:Class} \equivc{}~\hqd{class}; LogMapLt adds
\texttt{rdfs:subClassOf} \equivc{}~\hqd{causes\_by\_class}).

Second, \textbf{the discovery problem on this pair is genuinely hard, which is
the point of offering it as a test case}: the backbone correspondences are
exactly the ones lexical evidence cannot find. LogMap recovers 3 of 10; on
IES--BFO it recovers none, its only substantive mapping being
\ies{Entity}~\equivc{}~\bfo{entity}, the same bare root-label match both
LLMs produce in \S\ref{sec:llm}; even two mappings suffice for two
conservativity violations (\ies{ObjectName} \sub{} \ies{Entity}: a name becomes
an entity). Matching from scratch is harder than certifying an expert
crosswalk, and we do not claim to out-match matchers; the like-for-like
comparison is repair, next.

Third, \textbf{LogMap's internal repair is incomplete by design}: a scalable
propositional projection rather than full DL
reasoning~\citep{jimenez2011logmap}. This pair shows the residue concretely,
in the 87 new unsatisfiable classes that survive it. That is not a defect of
LogMap so much as a measurement of exactly the gap a complete reasoner in the
certification loop closes, a gap the OAEI's DISO track design already
anticipates by shipping repaired and unrepaired reference alignments side by
side.

\subsection{LogMap's repair facility versus orientation search}
\label{sec:repaircmp}

The like-for-like comparison: we gave LogMap's DEBUGGER facility the same
input orientation search received (the ten expert class correspondences
promoted to equivalences, over the complete IES and HQDM), with module
extraction and its post-repair HermiT check enabled.

The outcome is instructive on both sides. LogMap keeps all ten mappings and its
output is \emph{certifiably safe}: our harness confirms 0 new unsatisfiable
classes and 0 conservativity violations. Its repair, however, achieves safety by
a single uniform move: every equivalence is weakened to the same direction,
\hqd{X} \sub{} \ies{Y}, discarding the entire forward (IES-to-HQDM) content of
the crosswalk. Concretely, against the certified bridge of
Table~\ref{tab:bridge}:

\begin{itemize}
\item \textbf{All five provable equivalences are lost.} Orientation search
  proves \ies{Element} \equivc{} \hqd{spatio\_temporal\_extent} and four more;
  LogMap keeps only the reverse half of each.
\item \textbf{On the four divergent pairs, LogMap keeps the semantically wrong
  direction.} \hqd{individual} \sub{} \ies{Entity} asserts that every HQDM
  individual, including its activities, is an IES entity, precisely what
  IES denies by separating \ies{Event} from \ies{Entity}. The assertion is
  undetectable only because IES declares no disjointness axioms, and it
  survives S2 solely because LogMap simultaneously dropped the forward edges
  that would have exposed it. Orientation search, which preserves the expert's
  forward commitment as its baseline, rejects exactly this direction with a
  named counterexample.
\item \textbf{The mapping onto the broken class is kept, unflagged.}
  \hqd{participant} \sub{} \ies{EventParticipant} subsumes a provably empty
  class; coherence checking cannot distinguish vacuous from safe. Orientation
  search instead weakens the target one hop to \hqd{state} and keeps a
  contentful axiom.
\end{itemize}

The comparison is therefore not repair-fails-versus-repair-succeeds; both
outputs are coherent and conservative. It is \emph{retained logical content at
equal safety}: 10 uniformly-reversed axioms against 21 directed axioms with
five class and three property equivalences, the expert's forward direction
wherever safe, and a counterexample attached to every refusal. A criterion
that checks only coherence stops at the first safe answer; adding
conservativity and completion-toward-equivalence finds the strongest one.

\subsection{An LLM oracle}
\label{sec:llm}

LLM-as-oracle is now a standard component of matching systems, used to adjudicate
candidate correspondences. We wanted a direct test on this alignment, using our
certified bridge as gold, in the shape of the DISO-OAEI ranking task: given a
source class and a candidate pool from the target ontology, pick the best target
or return NIL.

The benchmark is 12 queries (8 into HQDM, 4 into BFO) built from the certified
bridge. We evaluate two systems: \texttt{Qwen3-Coder-30B-A3B-Instruct} at 8-bit,
and an IES-specialised fine-tune of the same base model, both served locally at
temperature 0. Results in Table~\ref{tab:llm}.

\begin{table}[h]
\centering
\caption{Mapping-oracle accuracy against the HermiT-certified bridge. $n=12$.}
\label{tab:llm}
\begin{tabular}{lccc}
\toprule
System & into HQDM (8) & into BFO (4) & total \\
\midrule
Qwen3-Coder-30B-A3B-Instruct (8-bit) & 5 & 1 & \textbf{6/12} \\
\quad + IES fine-tune                & 6 & 1 & \textbf{7/12} \\
\bottomrule
\end{tabular}
\end{table}

The aggregate is uninformative at this sample size and we do not claim a
significant difference between the two systems. The \emph{error structure} is
the result worth reporting, and it is consistent across both:

\begin{itemize}
\item \textbf{Both models pick \hqd{event} for \ies{Event}.} This is the false
  friend the crosswalk documents as its single most dangerous correspondence: a
  durative happening with participants mapped onto an instantaneous temporal
  boundary. The lexical pull of the identical label defeats both models, and the
  fine-tune does not help, because the trap is not a vocabulary problem.
\item \textbf{Neither model reaches \bfo{history}.} Asked for the BFO counterpart
  of \ies{Entity}, both answer \bfo{entity}: a shallow label match onto BFO's
  root. The certified answer requires knowing that a 4D whole-life individual is
  an occurrent, which is a commitment neither model retrieves.
\item \textbf{Both succeed where the mapping is easy.} \hqd{thing},
  \hqd{state}, \hqd{period\_of\_time}, \hqd{possible\_world} and
  \hqd{spatio\_temporal\_extent} are all recovered by both.
\end{itemize}

The pattern is that an LLM oracle is accurate exactly where a lexical baseline
would also be accurate, and fails exactly on the pairs where the two standards
genuinely diverge, which is where the alignment is hard and where the value
lies. On this evidence, an LLM is a reasonable candidate generator and a poor
adjudicator of upper-ontology disagreement; the reasoner is doing work the model
cannot substitute for. We report this as a probe with an explicit sample size,
not as a benchmark, and we say what would make it one in \S\ref{sec:limits}.

\section{Discussion}
\label{sec:discussion}

\paragraph{Publish the coherence report with the test case.} The 39-class
defect in \texttt{hqdm.owl} was invisible to everything except a DL reasoner
run on the input alone; it cost nothing to find and changes how any alignment
into HQDM should be read. A matching test case should ship a coherence report
on each input, and a reference alignment should flag any correspondence whose
target is unsatisfiable: precision and recall over such correspondences
measure agreement about the empty set.

\paragraph{Direction is information, and alignment formats should carry it.} A
correspondence published as $\langle s, o, 0.85 \rangle$ discards the
distinction between ``these are the same'' and ``every $s$ is an $o$ but not
conversely, and here is the individual that shows it.'' The second is more
useful and, as \S\ref{sec:bridge} shows, often the true relation between
standards built by different communities. SSSOM and EDOAL can express directed
correspondences; what is missing is the habit of computing the direction rather
than defaulting to equivalence. Orientation search is one way to compute it.

\paragraph{Relevance to OAEI.} The DISO track introduced in the OAEI 2026
campaign evaluates alignment over defence, intelligence and security ontologies,
and ships both a repaired and an unrepaired reference alignment
($R^{+}_{\approx}$ and $R_{\approx}$), which makes coherence an explicit
first-class concern of the evaluation rather than a post-processing step. This
paper's material has exactly that two-reference structure, produced by the
method itself: the naively promoted crosswalk of \S\ref{sec:naive} is an
unrepaired reference (consistent, incoherent, 218 conservativity violations)
and the certified bridge of \S\ref{sec:bridge} is its repaired counterpart,
with the delta between them fully explained by machine-found counterexamples.
The DISO collection ships IES, BFO and JC3IEDM but not HQDM, so the IES--HQDM
leg is exactly the join the collection cannot currently evaluate, and the pair
has a profile a track wants: small enough for exhaustive reasoner-based
evaluation, yet adversarial for lexical matchers, with an exact-label false
friend as its hardest correspondence and a correct BFO target sharing no
vocabulary with its source (\S\ref{sec:matchers} measures how adversarial;
DISO distributes IES~v5, so a contributed case would pin its lineage as
\S\ref{sec:background} does). We offer the three ontologies, the expert
crosswalk with its six documented divergences, and both reference alignments as
a candidate test case with one property the current track does not yet score:
the repaired reference is \emph{directed}, each non-equivalence carrying
its justifying counterexample, so a system can be scored not only on which
pairs it finds but on whether it gets the logical strength of each pair right.

\section{Limitations}
\label{sec:limits}

The crosswalk is a 17-correspondence backbone over two upper ontologies, not a
full alignment; above the 4D backbone the powertype hierarchies of the two
standards stop being parallel and become a genuine matching problem that we have
not attempted here. The correspondences were hand-curated by a single author and
their confidences are the curator's subjective strength rather than a measured
score; what is certified is the \emph{logical form} of each correspondence, not
its correctness as a claim about the domain. Orientation search's completion
phase is greedy under a fixed correspondence order, so the returned bridge is
maximal rather than provably maximum, and P1's weakening walks a single ancestor
chain rather than searching the full weakening lattice.

The most important limitation is one the BFO ablation makes concrete. Weakening
is confined to the ancestors of the asserted target, so the operator can never
reach a correct target that lies on a different branch. \S\ref{sec:bfo} is
exactly that case: from \bfo{material~entity} the ancestor chain leads only to
\bfo{entity}, where the axiom is safe but says nothing, and \bfo{history} is
unreachable. We supplied the four BFO targets by hand and used the search to
certify and orient them. Making re-targeting part of the search, rather than
part of the curator's judgement, is the obvious next step and we have not taken
it; a candidate generator over the target ontology, filtered by the same two
criteria, is the natural shape.

The systems comparison has its own fairness bounds: the July-2021 LogMap
standalone in default configuration (tuned or interactive use might do better);
the DEBUGGER fed the ten class correspondences only, not the three property
mappings; AgreementMakerLight~\citep{faria2013} and BERTMap~\citep{bertmap}
not run (older-JVM release line; per-pair language-model fine-tuning).
Extending to them, and to the conservativity-repair algorithms
of~\citep{solimando2014}, is the natural next experiment.

The LLM probe is 12 queries on two models from one family, which supports the
qualitative claim about error structure and supports no quantitative claim at
all; a proper version would cover multiple model families, multiple prompt
formats, and a candidate pool large enough for Hits@$k$ to be meaningful.
Finally, we certify against HermiT alone. Every result was re-checked by
invoking the HermiT command-line tool directly on the serialised merges,
bypassing owlready2's object model, and it agrees exactly (IES alone 0, HQDM
alone 39, IES+HQDM with the certified bridge 39 and the identical set, IES+BFO
with the certified bridge 0). That is a check on the harness, not an independent
prover; agreement with a second reasoner would strengthen the certificates.

\section{Related work}
\label{sec:related}

Alignment incoherence and its repair are the direct ancestors of this work.
Meilicke's analysis of alignment incoherence and the diagnosis-based repair
tradition~\citep{meilicke2011,meilicke2009} established that a coherent alignment
is the object of interest and that repair is a minimal-hitting-set problem over
mappings. LogMap~\citep{jimenez2011logmap} made scalable logic-based repair a
practical component of a matching system: its repair module identifies mappings
implicated in unsatisfiability and discards or weakens confidence in them during
matching. The conservativity principle and its violation-repair
algorithms~\citep{solimando2014} supplied the second criterion we use, and the
theory of conservative extensions and safe ontology
reuse~\citep{cuencagrau2008} is where that criterion comes from. Our
contribution relative to this line is narrow and, we think, useful: repair here
happens \emph{after} an expert has committed to the correspondence set, so
removal would discard curated knowledge, and the free variable that remains is
the \emph{direction} of each correspondence. Making direction the repair
variable, with completion toward equivalence and a counterexample attached to
every refusal, turns the repair log into documentation of where two government
standards disagree, arguably the more valuable artifact.

Axiom weakening~\citep{troquard2018} is the other parent, and Li and
Lambrix~\citep{lambrix2022} developed weakening-and-completing for EL
ontologies; our P1+P2 applies that shape at the alignment layer, to the
mapping target rather than an axiom of either input.

On the machine-learning side, BERTMap~\citep{bertmap} established the pattern of
a learned matcher followed by a symbolic repair stage, and recent work applies
LLMs as candidate generators or oracles over ontology
matching~\citep{llms4om}. Our \S\ref{sec:llm} probe is a small negative
datapoint for the oracle role precisely where lexical evidence misleads.
MELT~\citep{melt} is the evaluation platform on which the OAEI 2026 campaign
runs, including the DISO track that motivates \S\ref{sec:discussion}.

On the ontologies themselves, HQDM derives from West's data-modelling
work~\citep{west2011} in the BORO tradition~\citep{partridge2005}; BFO is
described in~\citep{arp2015} and standardised as~\citep{iso21838}. A companion census of
public ontologies~\citep{rovai2026census} reported that neither IES nor HQDM
declares exclusion structure adequate for counterfactual reasoning; this paper
asks the complementary question of what their declared axioms do to each other
when joined.

On IES--BFO specifically, Bailey et al.~\citep{bailey2025comparing}, a team
that includes the original authors of IES, BORO and BFO, compare the two
frameworks through parallel modelling of competency-question scenarios and
propose graph-level mapping patterns, including a pattern relating an IES4 4D
individual to a BFO material entity \emph{together with its} \bfo{history}.
Our certified bridge sharpens that insight: \ies{Entity} subsumes safely under
\bfo{history} but not \bfo{material~entity} (Table~\ref{tab:bfo}); of their
pattern's two legs, exactly one survives as an axiom. They
explicitly defer axiomatic, provable alignment to future work; the present
paper is that step for the 4D backbone. We previously released a SKOS-level (not
reasoner-checked) sketch crosswalk from JC3IEDM to
IES~\citep{rovai2026jc3iedm}; this is, to our knowledge, the first
\emph{certified} alignment involving IES and the first public IES--HQDM
alignment. Euzenat and Shvaiko~\citep{euzenat2013} remains the reference
text.

\section{Conclusion}

Two upper ontologies from the same intellectual tradition, aligned by an
expert on 17 backbone correspondences, produce a merge that passes a
consistency check and kills a hundred operational classes; one of them was already
killing 39 of its own before anyone aligned anything. Both facts are cheap to find with a reasoner, invisible without one,
and absent from any similarity score.

The repair is not deletion. Treating the direction of each correspondence as
the search variable, with a prover refereeing coherence and conservativity at
every step, returns a bridge that keeps every expert mapping, states full equivalence
wherever the standards agree, and attaches a machine-found counterexample
wherever they do not. No system we tested recovers that
bridge: the matchers fall for the false friend, and the repair that succeeds
discards every provable equivalence. BFO's 3D core is crossable only through
the occurrent branch: a 4D entity is a continuant's history, not the
continuant.

\section*{Declaration on Generative AI}

During the preparation of this work, the author used Anthropic Claude in order
to: draft and revise text, assist literature triage, and prepare LaTeX. The
locally served Qwen3-Coder-30B models in Section~\ref{sec:llm} are systems
under evaluation, not writing tools. All experiments, reported numbers and
reasoner runs were designed, executed and verified by the author. After using
this tool/service, the author reviewed and edited the content as needed and
takes full responsibility for the publication's content.

\appendix
\section{Data and code availability}
\label{sec:availability}

The crosswalk (SSSOM and SKOS/PROV-O), divergences, SHACL shapes, orientation
search, the three certified bridges with full decision/counterexample logs, the
LogMap/LogMapLt outputs with their evaluation harness
(\texttt{reasoning/systems/}, \texttt{systems\_eval.py}), and the LLM benchmark
are released under CC-BY-4.0 at
\url{https://github.com/fabio-rovai/ies-hqdm-crosswalk}; everything reproduces
with \texttt{python reasoning/orientation\_search.py} given any JDK. The input
ontologies are public (IES: OGL v3; HQDM: Apache-2.0, Crown Copyright; BFO
2020: CC-BY). This work is independent and not affiliated with or endorsed by
Dstl, GCHQ, the IES Working Group or the NDTP.

\end{document}